\pdfoutput=1

\documentclass[11pt]{article}

\usepackage{EMNLP2022}

\usepackage{times}
\usepackage{latexsym}

\usepackage[T1]{fontenc}

\usepackage[utf8]{inputenc}

\usepackage{microtype}

\usepackage{inconsolata}

\usepackage{color}

\usepackage{amsmath}
\usepackage{amssymb}
\usepackage{amsfonts}
\usepackage{booktabs}
\usepackage{graphicx}
\usepackage{multirow}
\usepackage{makecell}
\usepackage{setspace}
\usepackage{algorithm}
\usepackage{algorithmicx}
\usepackage{algpseudocode}

\graphicspath{{img/},}

\title{Know Where You're Going: \\ Meta-Learning for Parameter-Efficient Fine-Tuning}

\author{Mozhdeh Gheini, Xuezhe Ma, Jonathan May \\
  Information Sciences Institute \\
  University of Southern California \\
  \texttt{\{gheini, xuezhema, jonmay\}@isi.edu}}

\begin{document}
\maketitle
\begin{abstract}
A recent family of techniques, dubbed  lightweight fine-tuning methods, facilitates parameter-efficient transfer learning by updating only a small set of additional parameters while keeping the parameters of the pretrained language model frozen. While proven to be an effective method, there are no existing studies on if and how such knowledge of the downstream fine-tuning approach should affect the pretraining stage.
In this work, we show that taking the ultimate choice of fine-tuning method into consideration boosts the performance of parameter-efficient fine-tuning. By relying on optimization-based meta-learning using MAML with certain modifications for our distinct purpose, we \textit{prime} the pretrained model specifically for parameter-efficient fine-tuning, resulting in gains of up to 1.7 points on cross-lingual NER fine-tuning. Our ablation settings and analyses further reveal that the tweaks we introduce in MAML are crucial for the attained gains.
\end{abstract}

\section{Introduction}
\label{sec:intro}

The pretraining $\rightarrow$ fine-tuning paradigm, where a pretrained language model (PLM) is fine-tuned for each individual task, is the dominant practice in natural language processing, owing to state-of-the-art performance on a wide variety of tasks \cite{DBLP:journals/corr/abs-2003-08271}. The impressive effectiveness of this approach does not come at a low price. It requires iterative adjustment of anywhere between millions \cite{devlin-etal-2019-bert} to staggering billions of parameters \cite{DBLP:journals/corr/abs-2204-02311}. This upwards trend toward more parameters is likely to continue as it has been shown that parameter count directly correlates with fine-tuning effectiveness and final performance \cite{aghajanyan-etal-2021-intrinsic}. With this many parameters, fine-tuning \textit{all} parameters, as is common, becomes exceedingly computationally expensive: in settings where many models need to be fine-tuned, be it because of the sheer number of desired tasks or any case by case customization (e.g., training for several languages), serving a separate copy of all a model's parameters for each instance is costly in terms of storage.

\begin{figure}[t]
\centering
  \includegraphics[width=0.8\columnwidth]{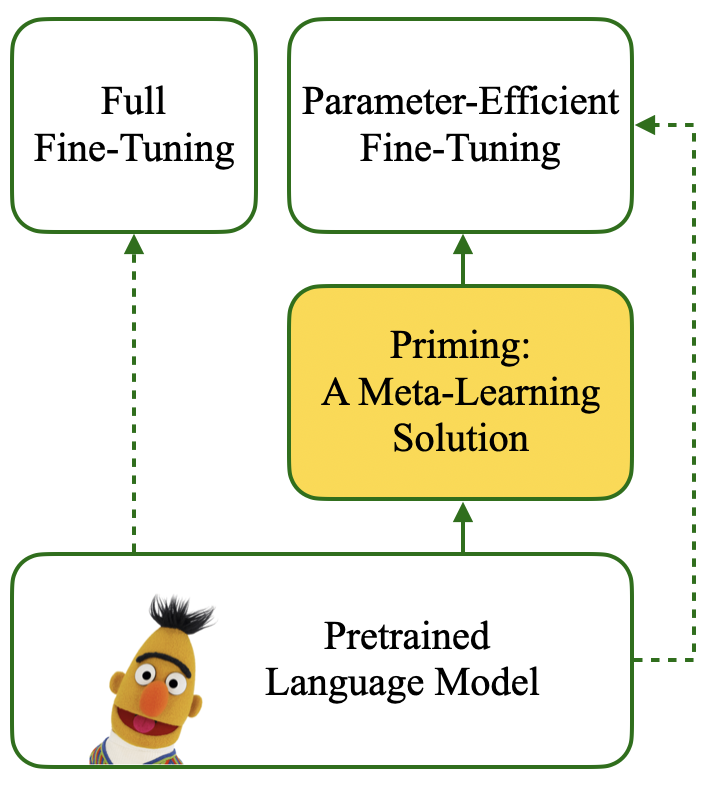}
\caption{Transfer learning for NLP pipeline; the shaded block is our contribution. Conventional transfer practice (dashed arrows) does not differentiate between full fine-tuning and parameter-efficient fine-tuning in any way. This work proposes a meta-learning solution to further modify and prime a pretrained model parameters to specifically target parameter-efficient fine-tuning.}
\label{fig:overview}
\end{figure}

Recent works on parameter-efficient (PE)\footnote{We use descriptors ``parameter-efficient'' and ``lightweight'' interchangeably.} fine-tuning address this issue by introducing methods that alternatively rely on only changing a tiny set of extra parameters \cite{houlsby2019parameter,li-liang-2021-prefix,hambardzumyan-etal-2021-warp,lester-etal-2021-power,hu2022lora,he2022towards} or a small fraction of the existing model's parameters \cite{DBLP:journals/corr/abs-2106-10199,gheini-etal-2021-cross}. These methods have been shown to be competitive with full fine-tuning despite modifying only as little as 0.01\% of all the parameters \cite{https://doi.org/10.48550/arxiv.2205.05638}.

With this shift in approach towards lightweight fine-tuning, we ask if the pretraining needs to be changed in any way as well. In other words, ought we modify the pretraining, knowing that we are going to opt for PE fine-tuning instead of full fine-tuning? Specifically, can we extend pretraining in a way that leads to \textit{parameter initializations that better suit PE fine-tuning} than the initializations coming outright from the PLM and used by full fine-tuning?

In this work, we show that, in fact, we can use optimization-based meta-learning to further modify the parameters from a PLM so that they are more beneficial for PE fine-tuning and result in improved performance on the target task after transfer. We term this step, which sits between conventional pretraining and fine-tuning, ``\textit{priming}'' (see Figure~\ref{fig:overview}). Specifically, as we describe in \S\ref{sec:algorithm}, we tweak the popular meta-learning approach MAML \cite{pmlr-v70-finn17a} for priming and crucially simulate the actual PE fine-tuning procedure in the inner loop of the algorithm. This means that instead of including all the parameters in the inner loop gradient update, we only consider those that will be updated by the lightweight fine-tuning method. Thus, during the meta-gradient update in the outer loop of the algorithm, this information about the ultimate fine-tuning approach will be incorporated into the pretrained values.

We choose cross-lingual transfer for named entity recognition (NER) as the testbed to show the effectiveness of our proposed priming stage. We show that priming a PLM boosts the performance of cross-lingual PE fine-tuning for NER by up to 1.7 F1 points. We provide the details of our lightweight fine-tuning setup in \S\ref{sec:exp}. Our ablation study in \S\ref{sec:abl1} reveals that simulating the fine-tuning procedure is indispensable to the observed improvements: it is not meta-learning in general, but \textit{how} we formulate the meta-learning setup that leads to observed gains.

Our \textbf{contributions} are:
\begin{itemize}
    \item We propose a meta-learning based mechanism termed ``priming'' to further update the parameters of a PLM in a way that improves the final PE transfer performance.
    \item We show the effectiveness of priming for cross-lingual transfer for NER as an exhibit.
    \item We justify and shed more light on the importance of the design elements in the priming algorithm through an ablation analysis.
\end{itemize}

\section{Meta-Learning Background}
\label{sec:background}

The meta-learning problem can be viewed as acquiring \textit{meta-parameters} $\theta$ using meta-training data $\mathcal{D}_{\text{meta-train}}$ such that $\theta$, when used for \textit{adaptation}, improves performance
on a new task with training data $\mathcal{D}_{\text{train}}$ \cite{stanford.cs330.fall2019.lec3}. Here, adaptation refers to any procedure that computes task parameters $\phi$ as a function of $\theta$ and $\mathcal{D}_{\text{train}}$: $\phi = f(\theta, \mathcal{D}_{\text{train}})$. While adaptation can be modeled in a multitude of ways, optimization-based meta-learning algorithms formulate the adaptation as an optimization procedure during which task parameters $\phi$ are obtained by fine-tuning the meta-parameters:

\begin{equation}
\label{eq:1}
    \phi = \theta - \alpha\nabla_\theta\mathcal{L}(\theta, \mathcal{D}_{\text{train}})
\end{equation}
where $\mathcal{L}$ is the task-dependent loss function.

Under this model of adaptation, meta-learning becomes a search for meta-parameters $\theta$ such that when used as initialization, optimal $\phi$ may be found via fine-tuning over many tasks. During meta-training, a ``task'' is modeled as a tuple of a training (\textit{support}) set $\mathcal{D}^{\text{tr}}$ and a testing (\textit{query}) set $\mathcal{D}^{\text{ts}}$. Hence, $\mathcal{D}_{\text{meta-train}} = \{(\mathcal{D}_1^{\text{tr}}, \mathcal{D}_1^{\text{ts}}), \cdots, (\mathcal{D}_n^{\text{tr}}, \mathcal{D}_n^{\text{ts}})\}$. Specifically, MAML \cite{pmlr-v70-finn17a}, which we take inspirations from, moves towards solution $\theta^\star$ for meta-parameters $\theta$ through a bi-level optimization procedure:

\begin{equation}
\label{eq:2}
    \theta^\star =
    \underbrace{
        \underset{\theta}{\arg\min}
            \sum_{\substack{(\mathcal{D}_i^{\text{tr}}, \mathcal{D}_i^{\text{ts}}) \\ \in \mathcal{D}_{\text{meta-train}}}}
                \mathcal{L}(
                    \overbrace{
                        \theta - \alpha\nabla_\theta\mathcal{L}(\theta, \mathcal{D}_i^{\text{tr}})
                    }^{\text{\textit{inner} optimization loop}},
                    \mathcal{D}_i^{\text{ts}}
                )
    }_{\text{\textit{outer} optimization loop}}
\end{equation}
where the inner loop takes gradient steps with respect to $\theta$ using the support set of each task to obtain task parameters $\phi_i$ for each one. The outer loop optimization process then takes meta-gradient steps with respect to $\theta$ by evaluating post-inner-update performance on the query set of each task, modifying $\theta$ to be a better initialization.

\section{Priming for Parameter-Efficient Fine-Tuning through Meta-Learning}
\label{sec:method}

Following the brief overview of optimization-based meta-learning in the previous section, we describe our proposed method. Our objective is to modify pretrained parameter values to be better aware of the downstream fine-tuning approach in order to lead to better final performance on the target task.

\begin{algorithm*}
\setstretch{1.08}
\caption{Priming for Lightweight Fine-Tuning (PE FT)}\label{alg:1}
\begin{algorithmic}[1]
\Require model $f_{\theta = \theta_p\cup\theta_h\cup\theta_a}$: pretrained params $\theta_p$, task head params $\theta_h$, and PE FT params $\theta_a$
\Require $\mathcal{D}_{\text{meta-train}} = \{(\mathcal{D}_1^{\text{tr}}, \mathcal{D}_1^{\text{ts}}), \cdots, (\mathcal{D}_n^{\text{tr}}, \mathcal{D}_n^{\text{ts}})\}$
\Require $\mathbb{L} = \{\mathcal{L}_1, ..., \mathcal{L}_t\}$: set of loss functions corresponding to all potential different tasks
\Require $\alpha$, $\beta$: learning rates
\Require $S$: number of inner gradient steps
\While{not converged}
    \State Sample a batch of tasks $\mathcal{T}$
    \ForAll{$\mathcal{T}_i \in \mathcal{T}$}
        \State $\theta^i = \theta$
        \For{$s\gets 1, \ldots, S$}
            \State $\theta_a^i = \theta_a^i - \alpha\nabla_{\theta_a^i}\mathcal{L}_{\mathcal{T}_i}(f_{\theta^i}, \mathcal{D}_{\mathcal{T}_i}^{\text{tr}})$; $\theta_h^i = \theta_h^i - \alpha\nabla_{\theta_h^i}\mathcal{L}_{\mathcal{T}_i}(f_{\theta^i}, \mathcal{D}_{\mathcal{T}_i}^{\text{tr}})$
    
            \hspace{9mm} \textcolor{red}{$\theta_p^i = \theta_p^i - \alpha\nabla_{\theta_p^i}\mathcal{L}_{\mathcal{T}_i}(f_{\theta^i}, \mathcal{D}_{\mathcal{T}_i}^{\text{tr}})$} \Comment{\textcolor{red}{In MAML, but not here as we are simulating PE FT.}}
        \EndFor
    \EndFor
    \State Meta-gradient steps $\theta_a = \theta_a - \beta\nabla_{\theta_a}\Sigma_{\mathcal{T}_i}\mathcal{L}_{\mathcal{T}_i}(f_{\theta^i}, \mathcal{D}_{\mathcal{T}_i}^{\text{ts}})$;

    \hspace{28.5mm} $\theta_p = \theta_p - \beta\nabla_{\theta_p}\Sigma_{\mathcal{T}_i}\mathcal{L}_{\mathcal{T}_i}(f_{\theta^i}, \mathcal{D}_{\mathcal{T}_i}^{\text{ts}})$
    \State $\theta_h = \theta_h^1$
\EndWhile
\State \Return $\theta_p$, $\theta_a$
\end{algorithmic}
\end{algorithm*}

\subsection{Problem Formulation}
Provided with a general-purpose PLM parameterized by a set of parameters $\theta_p$, and a dataset $\mathcal{D}$ for a target task, conventional fine-tuning practice for NLP is to add a \textit{task-specific head} parameterized by parameters $\theta_h$ (initialized randomly) to the PLM and update all parameters $\theta_p\cup\theta_h$ by taking gradient steps with respect to loss on $\mathcal{D}$. To avoid such expensive updates with an abundance of parameters, PE fine-tuning designates an additional set of parameters (initialized randomly) or only a small subset of $\theta_p$ as $\theta_a$ as the only parameters to be updated along $\theta_h$ while keeping $\theta_p \backslash \theta_a$ frozen.

With this alteration, perhaps prior to fine-tuning, $\theta_p$ can first be further updated to reach $\theta_p^\star$, which if transferred specifically under the parameter-efficient setting, results in better performance on the target task than $\theta_p$. We call this extra step between pretraining and fine-tuning and the problem of finding such parameters ``priming''. As an additional benefit, during priming we can also learn parameters $\theta_a^\star$ to be used instead of random initializations $\theta_a$ during PE fine-tuning. It is important to note that priming does not take away any of the benefits of lightweight fine-tuning: ultimately fine-tuning still relies on changing (and hence storing) the same number of parameters that would change without priming (which is equal to $|\theta_h| + |\theta_a^\star|$); it just starts from more suitable initialization points $\theta_p^\star$ and $\theta_a^\star$.

\subsection{Priming Algorithm}
\label{sec:algorithm}
We model priming as an optimization-based meta-learning problem. However, we refrain from directly applying MAML to it. This is due to the key observation that under PE fine-tuning, the adaptation procedure, as shown in Equation~\ref{eq:1}, has changed: only a subset of parameters are updated during adaptation. Hence, it should be properly simulated in the inner loop in Equation~\ref{eq:2}, and we need to adapt the MAML algorithm accordingly.

Algorithm~\ref{alg:1} outlines the adaptations used for priming. The inner loop (lines 3-8) simulates exactly how we are going to ultimately fine-tune in a lightweight fashion by only updating $\theta_a$ and $\theta_h$. Note that the statement marked as red and without a line number, which additionally updates pretrained parameters $\theta_p$, would be executed by MAML. But we crucially omit it in our proposed priming algorithm. At the end of the outer loop (line 9), we take meta-gradient steps with respect to the parameters the initializations of which we are trying to enhance, $\theta_a$ and $\theta_p$. As $\theta_h$ will be initialized from scratch for each new task at the time of fine-tuning, we do not compute meta-gradients for it, and simply assign it to one of the calculated sets in the inner loop, e.g., the first set corresponding to the first task in the sampled batch of tasks ($\theta_h = \theta_h^1$ on line 10).

\section{Experimental Setup}
\label{sec:exp}

While our proposed priming algorithm is model-agnostic, we need a concrete PE fine-tuning and meta-training setup for empirical evaluation.

\begin{figure}[t]
\centering
  \includegraphics[width=0.9\columnwidth]{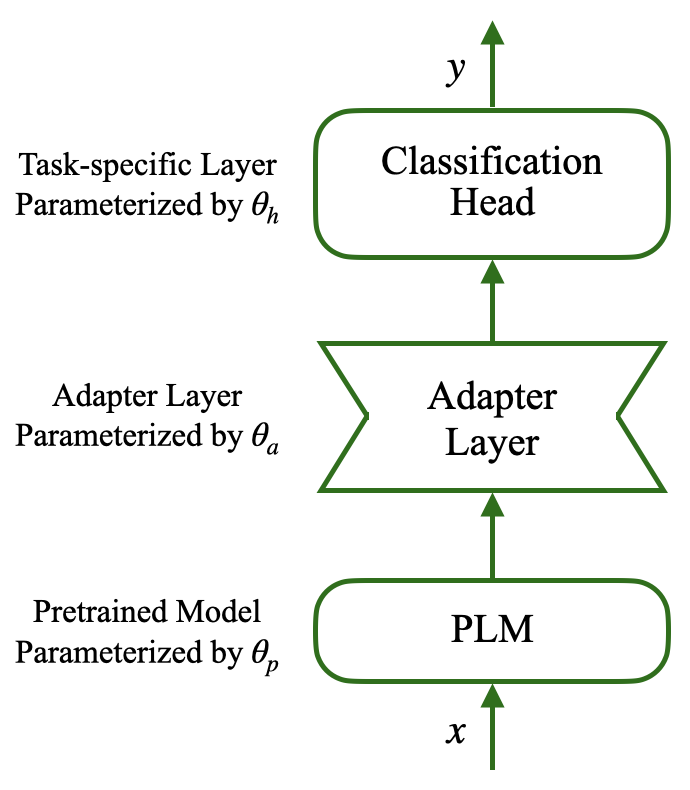}
\caption{Overall model architecture used in our experiments. $\theta_a$ comprises a single adapter layer directly after the pretrained model.}
\label{fig:arch}
\end{figure}

For lightweight fine-tuning, we choose adapters \cite{houlsby2019parameter}. Adapter modules comprise a sequence of a down-project transformation, a non-linearity, and an up-project transformation. For lightweight fine-tuning, they get inserted after the multi-headed attention, and the feedforward layer in each layer of a Transformer-based \cite{NIPS2017_3f5ee243} PLM and get updated during transfer. For our experiments, we only add a single adapter module after the last layer of the pretrained Transformer (for an illustration, see Figure~\ref{fig:arch}). Our model then computes the logits for input as:

\begin{center}
    $h(g(f(x; \theta_p); \theta_a); \theta_h)$
\end{center}
Where $f$ is the pretrained model, $g$ is the single adapter layer at the top, and $h$ is the task-specific head.

As a testbed, we experiment with cross-lingual NER. For this specific case, we can then design the priming (meta-learning) and fine-tuning stages as such:
\begin{itemize}
    \item Meta-Learning:  Using one or more source languages (each representing what a target task could look like), we construct the meta dataset and run Algorithm~\ref{alg:1}. Per our problem formulation, $\theta_p$ and $\theta_a$ are shared among all languages. But each source language $l$ will have its own separate head parameterized by $\theta_{h_l}$.
    \item Fine-Tuning: For each desired target language, we use the pretrained and adapter parameter initializations acquired during meta-learning along with randomly initialized new head parameters as the model's starting point. We then fine-tune only the adapter parameters and the head parameters. In our single adapter layer setup, this means only updating less than \textbf{0.4\%} of all the parameters.
\end{itemize}

\renewcommand{\arraystretch}{1.2}
\begin{table*}[t]
    \centering
    \scalebox{0.9}{
    \begin{tabular}{clrrrrrr}
        \toprule
         &  & \textbf{Hindi} & \textbf{Afrikaans} & \textbf{Azerbaijani} & \textbf{Lithuanian} & \textbf{Estonian} & \textbf{Dutch} \\
        \cmidrule(lr){3-8}
        \parbox[t]{8mm}{\multirow{3}{*}{\rotatebox[origin=c]{90}{\makecell{Without \\ Priming}}}} & \textbf{1/Full FT} (100\%) & \textbf{86.73} & \textbf{91.29} & \textbf{87.70} & \textbf{89.43} & \textbf{90.88} & \textbf{91.47} \\
         & \textbf{2/HT} (3e-3\%) & 72.71 & 79.11 & 74.24 & 78.34 & 81.23 & 78.90 \\
         & \textbf{3/AT} (0.4\%) & 77.76 & 84.10 & 81.08 & 83.00 & 85.13 & 83.89 \\
        \midrule
        \midrule
        \parbox[t]{8mm}{\multirow{4}{*}{\rotatebox[origin=c]{90}{\makecell{With \\ Priming}}}} & \textbf{4/Meta Priming $\rightarrow$ AT} & \textbf{81.25} & \textbf{88.23} & \textbf{83.29} &\textbf{86.30} & \textbf{87.18} & \textbf{88.97} \\
         & \textbf{5/FT Priming $\rightarrow$ AT} & 79.55 & 87.78 & 82.28 & 85.79 & 86.56 & 88.82 \\
         & \textbf{6/MP [MAML Loop] $\rightarrow$ AT} & 80.18 & 85.82 & 81.47 & 85.73 & 85.86 & 88.13 \\
         & \textbf{7/MP [1 Inner Step] $\rightarrow$ AT} & 80.23 & 86.46 & 80.35 & 84.43 & 86.14 & 88.49 \\
        \bottomrule
    \end{tabular}}
    \caption{Entity-level F1 under each of the fine-tuning settings for NER tasks across six languages. Bold numbers indicate top-scoring approaches in each category (with or without priming). Percentages in parentheses next to each setting are the fraction of parameters that need to be updated and ultimately stored (all \textbf{AT} settings have the same percentage). Priming as described in Algorithm~\ref{alg:1} is most effective in improving PE fine-tuning performance and closing the gap with full fine-tuning.}
    \label{tab:res}
\end{table*}

\subsection{Data Details}
\label{sec:data}
We use the WikiAnn multilingual NER dataset \cite{pan-etal-2017-cross}, which is in the Wikipedia domain and available from the Datasets Python library \cite{lhoest-etal-2021-datasets}. The train, validation, and test splits, as provided by \newcite{rahimi-etal-2019-massively}, range from 100 to 20k instances. In our experiments, we use the English and Spanish sets as source languages, each with 20k instances during the priming stage. At fine-tuning, we evaluate the quality of transfer for six target languages: Hindi (5k instances), Afrikaans (5k), Azerbaijani (10k), Lithuanian (10k), Estonian (15k), and Dutch (20k).

\subsection{Implementation Details}
\label{sec:impl}
Our implementation is based off of the Transformers \cite{wolf-etal-2020-transformers} and PyTorch Lightning \cite{Falcon_PyTorch_Lightning_2019} libraries. For our pretrained model, we use multilingual BERT (mBERT, \texttt{bert-base-multilingual-cased}) \cite{devlin-etal-2019-bert}. For the adapter layer, we set the bottleneck dimension as 64 in our experiments.

Regarding the details of the meta-training implementation, the meta-gradient at the end of the outer loop relies on gradients of gradients, which results in second-order gradients. As a workaround to avoid that, we use a first-order approximation (similar to first-order MAML \cite{pmlr-v70-finn17a}). We further discuss this decision in \S\ref{sec:abl2}. For the inner loop, we take five steps of stochastic gradient descent with a learning rate of 0.03. For the outer loop optimization, we use AdamW optimizer \cite{loshchilov2018decoupled} with a learning rate of 5e-5 and a linear learning rate scheduler.

\subsection{Baselines and Method Evaluation Settings}
\label{sec:settings}
With empirical evaluation design choices described, in this section, we outline the experimental settings. To assess the effectiveness of priming, we run two categories of experiments. First are the settings that include no priming at all:

\paragraph{1/Full fine-tuning baseline.} This corresponds to fine-tuning $\theta_p\cup\theta_h$ where $\theta_h$ is initialized randomly. It provides an upper bound/strong competition for what we can aim for with PE fine-tuning.

\paragraph{2/Head tuning baseline.} This corresponds to freezing $\theta_p$ (treating the pretrained model as a feature extractor) and fine-tuning $\theta_h$ where $\theta_h$ is initialized randomly. It provides a lower bound for PE fine-tuning.

\paragraph{3/Adapter tuning baseline.} This corresponds to fine-tuning $\theta_a\cup\theta_h$. It is the baseline PE fine-tuning, and we want to investigate if priming can help improve upon it. \\

Besides the settings above, we also experiment with the ones below, which incorporate priming:

\paragraph{4/Adapter tuning after priming as in Alg.~\ref{alg:1}.} This corresponds to fine-tuning $\theta_a\cup\theta_h$ where $\theta_p$ (frozen) and $\theta_a$ are acquired through running priming before fine-tuning, and $\theta_h$ is initialized randomly. Compared to the adapter tuning baseline (3), it measures how much priming can improve PE fine-tuning by incorporating such knowledge into parameters.

\paragraph{5/Adapter tuning after priming through fine-tuning.} This setting is the same as setting 4 except that instead of using Alg.~\ref{alg:1} for priming, we simply fine-tune $\theta_p\cup\theta_a\cup\theta_h$ on the same data that would have constructed the meta dataset before proceeding with PE fine-tuning just as in setting 4. This is to illustrate that mere exposure to data during priming is not enough, and treating it as an optimization-based meta-learning problem is beneficial. \\

Additionally, we have two ablation settings to study the effect of the presence or absence of the MAML-specific red line and the number of inner steps in Algorithm~\ref{alg:1}, which we will discuss in more detail in \S\ref{sec:abl1} and \S\ref{sec:abl2}.

\begin{figure*}[t]
\centering
  \includegraphics[width=\textwidth]{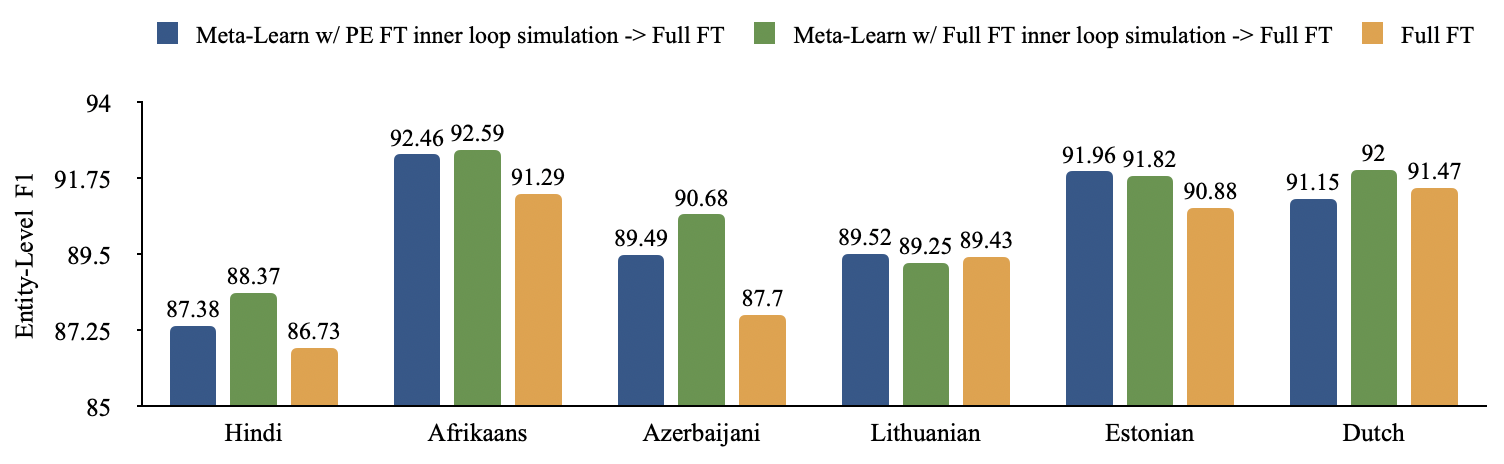}
\caption{Comparison between different priming strategies for downstream full fine-tuning. In this case, as opposed to parameter-efficient fine-tuning, it is beneficial to use full fine-tuning in the inner loop.}
\label{fig:chart}
\end{figure*}

\section{Results and Analysis}
\label{sec:res}

Our main results are reported in Table~\ref{tab:res}. All scores are entity-level micro F1. The setting numbers in the table match the setting description numbers in \S\ref{sec:settings} (e.g., \textbf{1/Full FT} $\leftrightarrow$ \textbf{1/Full fine-tuning baseline}). Among all PE fine-tuning settings without any priming and those with priming, \textbf{4/Meta Priming $\rightarrow$ AT}, which is the materialization of our proposed priming algorithm (Algorithm~\ref{alg:1}), is the best-performing approach. In comparison with baseline lightweight fine-tuning (\textbf{3/AT}), our approach results in gains of up to 5.08 points, indicating that priming with the knowledge of the ultimate transfer process is substantially helpful. Additionally, the approach results in gains of up to 1.7 points compared to fine-tuning-based priming (\textbf{5/FT Priming $\rightarrow$ AT}), signifying that it is not just a matter of exposure to more data, but a matter of appropriately using the extra exposure to simulate the eventual fine-tuning approach.

\subsection{Ablation Setting 1: Substitute MAML Inner Loop}
\label{sec:abl1}
To highlight the importance of the change we introduce in MAML for our purposes (removal of the red statement in Algorithm~\ref{alg:1}), we run an ablation setting \textbf{6/MP [MAML Loop] $\rightarrow$ AT} that is essentially \textbf{4/Meta Priming $\rightarrow$ AT} where we additionally execute the red statement and consequently update all parameters in the inner loop.

It can be observed across the board that, in fact, omitting the red statement, which is essentially simulating the downstream PE fine-tuning setting, is essential for superior performance.

As a matter of fact, we can generalize the question asked at the core of this work. Beyond parameter-efficient fine-tuning, can we expect gains by using optimization-based meta-learning and simulating the eventual fine-tuning strategy, whatever it might be? If so, while including the red statement hurts downstream parameter-efficient fine-tuning, it should improve the performance of full fine-tuning.

To confirm this, we repeat the settings in this section (\textbf{4/Meta Priming $\rightarrow$ AT} and \textbf{6/MP [MAML Loop] $\rightarrow$ AT}), replacing adapter tuning (\textbf{AT}) with full fine-tuning. The results are provided in Figure~\ref{fig:chart}. As expected, in most cases (all but Lithuanian and Estonian) including the red statement and making the inner loop simulate full fine-tuning (MAML inner loop) results in superior performance for downstream full fine-tuning (green bar in the middle in each series) compared to PE inner loop simulation. We hypothesize that the discrepancy in the case of Lithuanian and Estonian is due to the fact that full fine-tuning is very powerful in general, and potentially more robust to heterogeneous priming conditions. Additionally, it is satisfactory to observe that priming provides gains on top of full fine-tuning without any priming at all (blue and green bars are higher than the yellow bar on the right in each series).

\begin{figure}[ht]
\centering
  \includegraphics[width=0.85\columnwidth]{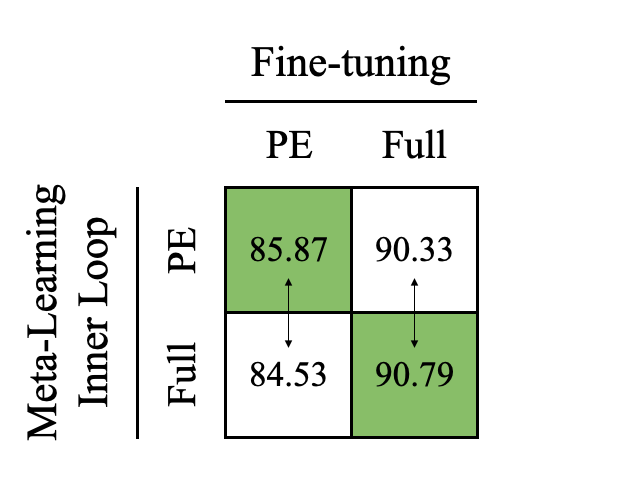}
\caption{Know Where You're Going: Best performances appear on the diagonal, where a homogeneous priming strategy is devised before downstream fine-tuning. Directly comparable blocks are linked using arrows.}
\label{fig:matrix}
\end{figure}

Figure~\ref{fig:matrix} provides an overview of what we recommend based on our experiments. It displays all four possible combinations of priming strategy $\rightarrow$ ultimate fine-tuning strategy sequences. Each block reports the average performance of downstream fine-tuning for NER across the six languages in our experiments using the corresponding combination. The best performing in each case happens on the main diagonal of the matrix. Therefore, for best performance, ideally, a priming stage simulating the subsequent fine-tuning strategy should be included in the transfer pipeline.

Additionally, a point worth noting from an engineering perspective for working with different priming strategies is that during our experiments, we found training the default MAML inner loop, which includes all the parameters, to
require a lower learning rate for proper convergence than the recommended inner loop in our priming algorithm (learning rate of 1e-4 vs. 0.03).

\subsection{Ablation Setting 2: Number of Inner Steps}
\label{sec:abl2}
\begin{figure}[ht]
\centering
  \includegraphics[width=0.5\columnwidth]{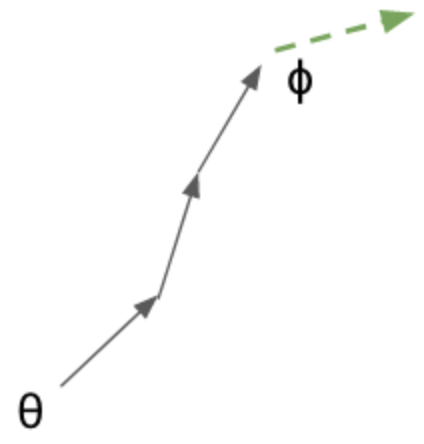}
\caption{First-order approximation of meta-gradient update. Illustration courtesy of \newcite{fomaml_tds_post}. After three steps of inner updates ($\theta\rightarrow\rightarrow\rightarrow\phi$), $\theta$ is updated by the gradient of the query set loss evaluated at $\phi$ (the green dashed arrow).}
\label{fig:fomaml}
\end{figure}

As mentioned in \S\ref{sec:impl}, we use a first-order approximation of the meta-gradients. A visualization of what that means in terms of actual parameter updates by \newcite{fomaml_tds_post} is provided in Figure~\ref{fig:fomaml}. Meta-parameters $\theta$ are updated in the direction of the gradient of the query set loss calculated at the value reached at the end of the inner loop. Hence, the fewer the number of inner steps, the more the updates will be similar to those under regular fine-tuning (in the limit of zero inner steps, it will be equivalent to conventional fine-tuning). The number of inner steps is critical for reaching better initialization. To highlight that, we also experiment with an ablation setting \textbf{7/MP [1 Inner Step] $\rightarrow$ AT} that is identical to \textbf{4/Meta Priming $\rightarrow$ AT} with only one inner step.

Indeed, the importance of the number of inner steps is reflected in the performance gap between the two settings, with \textbf{4/Meta Priming $\rightarrow$ AT} always being better.

\section{Related Work}

Our work takes inspiration from and can be contextualized within both the existing lightweight fine-tuning literature and meta-training literature. Lightweight fine-tuning methods are a response to the ever-growing size of the PLMs, which makes full fine-tuning prohibitively expensive. Recently, different flavors of PE fine-tuning have been explored. One category includes methods that add and solely update a new set of parameters; like adapters \cite{houlsby2019parameter}, prefix tuning \cite{li-liang-2021-prefix}, and LoRA \cite{hu2022lora}, to name a few. Another category of methods does not add any additional parameters and instead relies on updating a small subset of existing parameters of the pretrained model; for instance, BitFit \cite{DBLP:journals/corr/abs-2106-10199} and exclusive cross-attention fine-tuning \cite{gheini-etal-2021-cross}.

Despite the rich literature on different parameter-efficient transfer approaches, to the best of our knowledge, no existing study investigates whether in response pretraining practices need to be updated in any way. In this work, we attempt to address that void. \newcite{he2022towards} provide a unified framework within which several flavors of lightweight fine-tuning can be interpreted. Therefore we, while studying an adapter-based approach in this work, expect priming to be fundamentally applicable and useful to other flavors too.

We are also inspired by the body of work that takes advantage of optimization-based meta-learning to come by initializations that would be better suited for a specific objective. \newcite{xia-etal-2021-metaxl} use meta-learning to learn representation transformations that transform representations of a high-resource language in a way that they become more beneficial for effective transfer to low-resource languages.  \newcite{nooralahzadeh-etal-2020-zero} effectively use meta-learning to leverage training data for zero-shot and few-shot cross-lingual transfer on Question Answering and Natural Language Inference. \newcite{NEURIPS2019_f4dd765c} use a meta-objective to optimize representations for continual learning.

Perhaps closest in spirit to our objective and trying to bring these two lines of work together, \newcite{DBLP:journals/corr/abs-2110-15943} offer a meta-learning-like solution to ``\textit{learn to learn in context}'': using our terminology, while we address priming for PE fine-tuning, they address priming for in-context learning \cite{NEURIPS2020_1457c0d6}. In-context learning is a few-shot learning technique with no additional training required, where an LM is used to label new instances after conditioning on only a few supervised examples. \newcite{DBLP:journals/corr/abs-2110-15943} propose to better prepare the model for such an inference process on a new unseen task by including a tuning stage where the model is trained to do the same on simulated input sequences from a set of available tasks. The extra training stage that they include can be seen as equivalent to our priming stage, where in both cases, the goal is to prepare the model for what is subsequently coming.

\section{Conclusion}

We show that modifications to the fine-tuning procedure can have the most positive impact when accompanied by respective modifications to the pretraining procedure. Specifically, in the case of parameter-efficient fine-tuning, we propose to add an extra ``priming'' step between the conventional pretraining and fine-tuning steps to incorporate awareness of the ultimate transfer procedure in the pretrained values. We model this desideratum as an optimization-based meta-learning problem, which integrates such knowledge by means of further updating pretrained parameters under a simulation PE fine-tuning. We show the effectiveness of our priming method in improving baseline lightweight fine-tuning on cross-lingual transfer for NER. Through further analysis, it is revealed that both our decisions to 1) model priming with meta-learning instead of simple fine-tuning and 2) simulate the actual PE fine-tuning in the meta-learning instead of using it unadjusted contribute to the effectiveness of priming.

\bibliography{anthology,custom}
\bibliographystyle{acl_natbib}

\appendix


\end{document}